\documentclass[lettersize,journal]{IEEEtran}
\usepackage{amsmath,amsfonts}
\usepackage{algorithmic}
\usepackage{algorithm}
\usepackage{array}
\usepackage[caption=false,font=normalsize,labelfont=sf,textfont=sf]{subfig}
\usepackage{textcomp}
\usepackage{stfloats}
\usepackage{url}
\usepackage{verbatim}
\usepackage{graphicx}
\usepackage{cite}
\usepackage{booktabs}
\begin{document}

\title{Road Disease Detection based on Latent Domain Background Feature Separation and Suppression}

\author{
First A. JUWU ZHENG, Second B. JIANGTAO REN
\thanks{This work was supported in part by the Key R\&D projects of GuangDong Province under
Grant 2022B0101070002.\emph{ Corresponging author: Jiangtao Ren.}\\  \indent First A. JUWU ZHENG 
is with the School of Computer Science and Engineering, Sun Yat-Sen University, Guangdong, 510275 
China (e-mail: zhengjuwu29@mail2.sysu.edu.cn).\\ \indent Second B. JIANGTAO REN is with the 
School of Computer Science and Engineering, Sun Yat-Sen University, Guangdong, 510275 
China (e-mail: issrjt@mail.sysu.edu.cn).}
}
\markboth{The paper headers}%
{Shell \MakeLowercase{\textit{et al.}}: Road Disease Detection based on Latent Domain Background Feature Separation and Suppression}
\IEEEpubid{}
\maketitle
\begin{abstract}
  Road disease detection is challenging due to the the small proportion of 
  road damage in target region and the diverse background,which
  introduce lots of domain information.Besides, disease categories have 
  high similarity,makes the detection more difficult. In this paper, we propose 
  a new LDBFSS(Latent Domain Background Feature Separation and Suppression)
  network which could perform background information 
  separation and suppression without domain supervision and contrastive enhancement of 
  object features.We combine our LDBFSS network with YOLOv5 model to enhance disease features
  for better road disease detection.
  As the components of LDBFSS network, we first design a latent domain discovery module and a domain adversarial 
  learning module to obtain pseudo domain labels through unsupervised method, 
  guiding domain discriminator and model to train adversarially to suppress 
  background information. In addition, we introduce a contrastive 
  learning module and design k-instance contrastive loss, optimize the disease 
  feature representation by increasing the inter-class 
  distance and reducing the intra-class distance for object features.
  We conducted experiments on two road disease detection datasets,
  GRDDC and CNRDD, and compared with other models,which show an increase of nearly 4\%
  on GRDDC dataset compared with optimal model, and an
  increase of 4.6\% on CNRDD dataset.
  Experimental results prove the effectiveness and superiority
  of our model. 
\end{abstract}


\section{INTRODUCTION}
\IEEEPARstart{A}{s} a common public facility, roads are damaged to varying degrees under the influence 
of long-term weather, external forces and other factors, resulting in  various road surface 
diseases and potential traffic safety hazards. It is of great help to prevent the further 
deterioration of road conditions and prevent various traffic safety problems by timely 
detection and maintenance of road diseases. Therefore, all countries attach great importance 
to road condition assessment \cite{ref1}. The traditional pavement disease detection is completed 
manually, which requires a lot of manpower and material resources, and it is easy to miss the 
disease due to the fatigue of the inspector. With the development of deep learning in recent years
, we can fully tap the potential of deep learning technology 
in the field of intelligent transportation applications, such as automatic driving \cite{ref2,ref3,ref4,ref5}, 
traffic analysis \cite{ref6,ref7}, traffic accident detection \cite{ref8,ref9},road disease detection 
and so on.
\par
\IEEEpubidadjcol
Road disease detection methods can be divided into two categories, image processing-based 
methods and deep network-based methods.The method based on image processing uses the 
representative imaging features of diseases 
to detect, such as Gabor filter, directional gradient(HOG), local binary pattern(LBP) 
\cite{ref10,ref11,ref12,ref13}. Although these methods have achieved certain results on relatively simple datasets,
they are often limited when facing more 
complex road conditions and detection objects. The method based on deep learning can be 
competent for more complex disease detection tasks by using the powerful modeling and feature 
extraction capabilities of deep networks. Pavement disease detection based on deep learning 
can be divided into disease semantic segmentation and disease target detection. The former has 
high cost and poor real-time performance due to the need for pixel-level labeling and 
prediction, so disease target detection has become the mainstream of road disease detection. 
\cite{ref14} used multi-scale features and various enhancement methods to improve the performance 
of disease recognition.In \cite{ref15} the author explored the different effects of various backbone 
networks on road disease recognition. \cite{ref16} A two-stage inspection method was used to detect
concrete roadside cracks, and the effects of different light and weather conditions on the 
detection were studied. \cite{ref17} Explored a new data enhancement method, and generated new 
road damage training data by generative adversarial network and Poisson hybrid method.
\cite{ref18} introduced coordinate attention to 
improve positioning accuracy, and introduced K-means clustering to improve classification 
accuracy. \cite{ref19} Using adversarial learning to improve the performance of the model in 
domain adaptive scene. \cite{ref20} combined with image processing methods to provide 
prior information for the target detection model, improved the detection performance of 
the model.
\begin{figure}[htbp]
  \begin{minipage}[t]{0.25\linewidth}
      \centering
      \includegraphics[height=1.0\textwidth,width=\textwidth]{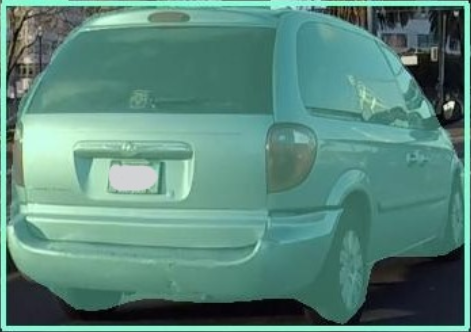}
      \centerline{(a)}
  \end{minipage}%
  \begin{minipage}[t]{0.25\linewidth}
      \centering
      \includegraphics[height=1.0\textwidth,width=\textwidth]{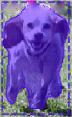}
      \centerline{(b)}
  \end{minipage}%
  \begin{minipage}[t]{0.25\linewidth}
      \centering
      \includegraphics[height=1.0\textwidth,width=\textwidth]{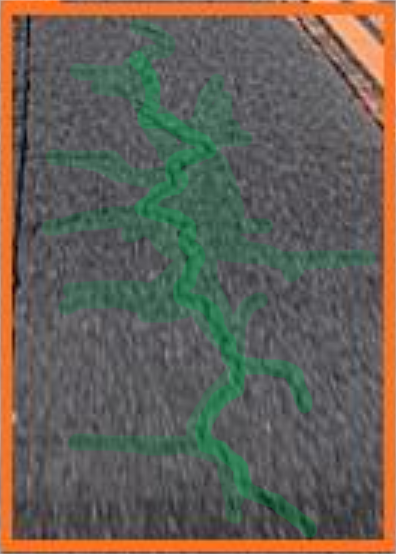}
      \centerline{(c)}
  \end{minipage}%
  \begin{minipage}[t]{0.25\linewidth}
      \centering
      \includegraphics[height=1.0\textwidth,width=\textwidth]{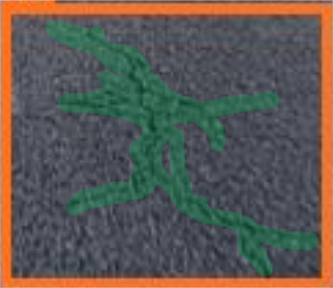}
      \centerline{(d)}
  \end{minipage}
  \caption{The road damage category has a smaller region proportion.
  (a)(b)common detection classes.(c)(d)disease detection classes.}
  \label{fig_introduction_2}
\end{figure}
\begin{figure}[htbp]
  \begin{minipage}[t]{0.33\linewidth}
      \centering
      \includegraphics[width=\textwidth]{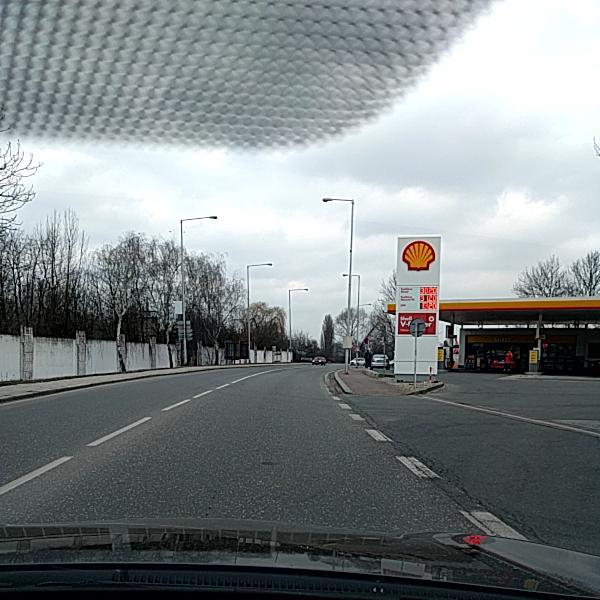}
      \centerline{(a)}
  \end{minipage}%
  \begin{minipage}[t]{0.33\linewidth}
      \centering
      \includegraphics[width=\textwidth]{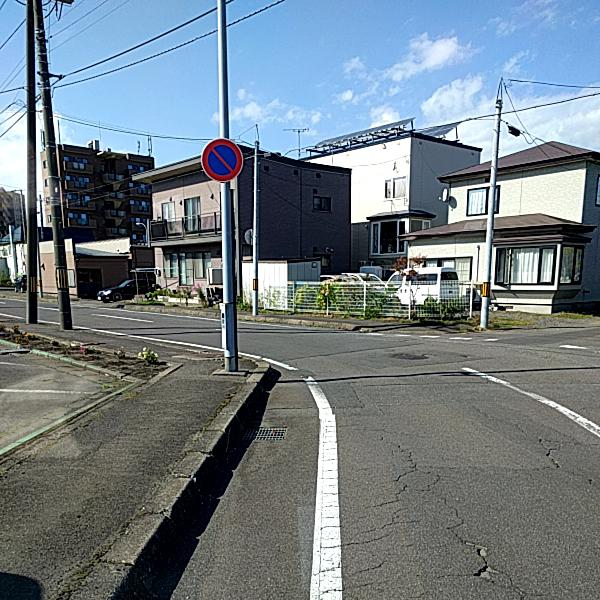}
      \centerline{(b)}
  \end{minipage}%
  \begin{minipage}[t]{0.33\linewidth}
      \centering
      \includegraphics[width=\textwidth]{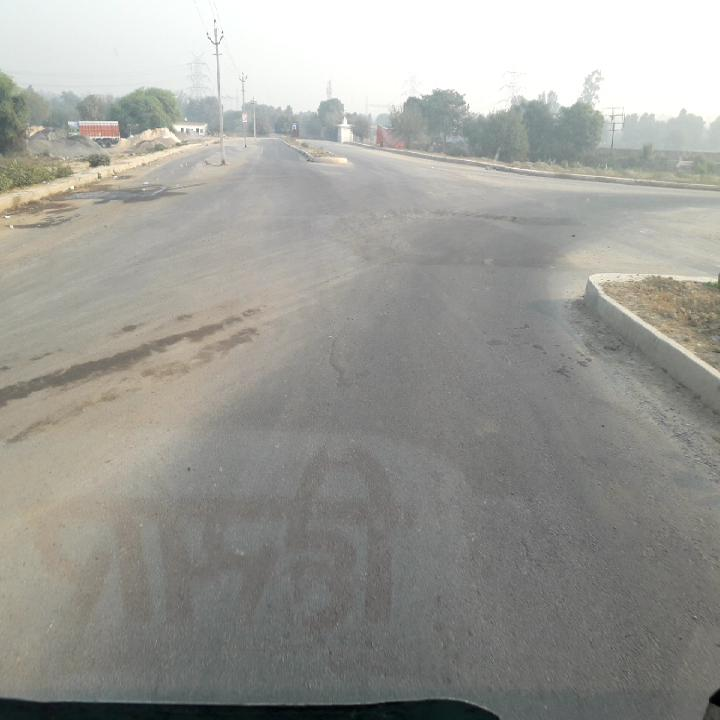}
      \centerline{(c)}
  \end{minipage}
  \caption{Domain gap between images from different country.(a)Czech.(b)Japan.(c)India.}
  \label{fig_introduction_3}
\end{figure}
\par
We notice that in common target detection task, the target has a relatively large 
proportion of annotation region, while the road disease target area only accounts for a small 
part, thus introduce lots of background information, as shown in Fig.\ref{fig_introduction_2}. Besides,the sources of road pictures are
 different, making the background diverse,as shown in Fig.\ref{fig_introduction_3}.
Background information in road disease detection poses two problems. One is that for the same disease category, 
diverse background information will prevent us from obtaining a unified disease representation. 
The second is that there may be similar background information between different categories, which confuses 
the model's distinction of diseases.We formulate this as a domain generalization problem in 
road disease detection
and conduct research on domain adaptation 
and domain generalization. \cite{ref42} used 
adversarial learning and metric learning to achieve domain adaptation, \cite{ref43} considered the 
influence of noise features unrelated to the target domain on domain alignment, and purified the 
features. \cite{ref44} studied the domain adaptation problem on multimodality. \cite{ref45} used domain 
labels to train domain classifiers and used this training model to obtain image representations 
without domain information. In \cite{ref46}, the author used multiple pre-training models and used 
ensemble learning methods to obtain the optimal results.
\par
\begin{figure}[htbp]
  \begin{minipage}[t]{0.5\linewidth}
      \centering
      \includegraphics[width=\textwidth]{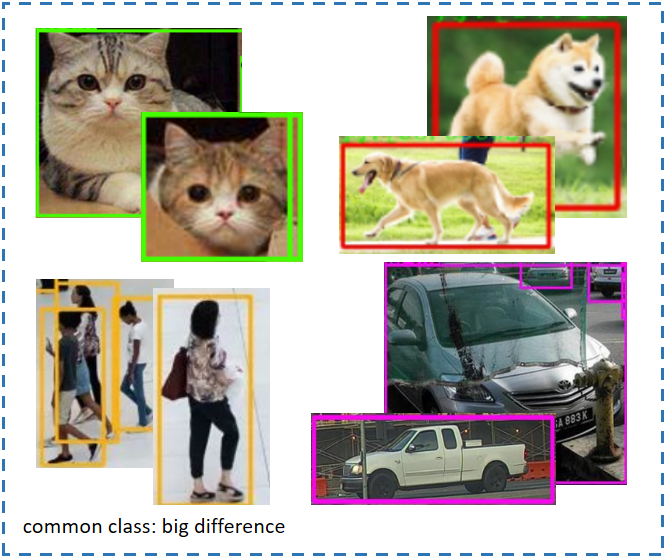}
      \centerline{(a)}
  \end{minipage}%
  \begin{minipage}[t]{0.5\linewidth}
      \centering
      \includegraphics[width=\textwidth]{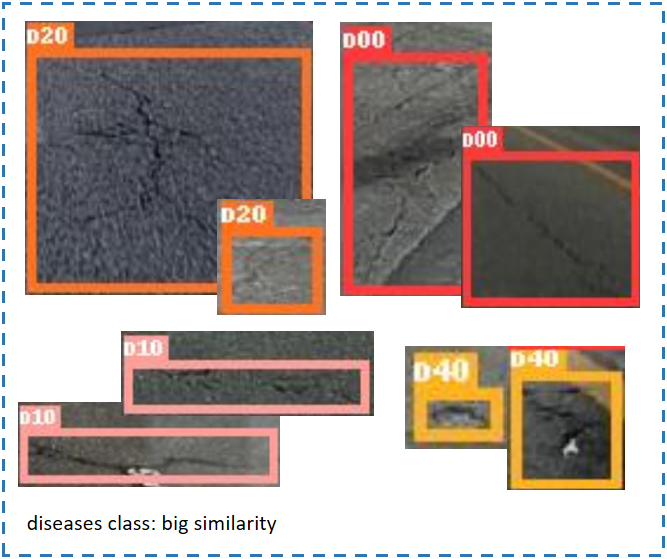}
      \centerline{(b)}
  \end{minipage}
  \caption{Target in disease detection have greater similarity.
  (a)common detection classes such as cars,dogs,people and cars.(b)disease detection classes 
  such as longitudinal crack,transverse crack, aligator crack and pothole.}
  \label{fig_introduction_1}
\end{figure}
In addition, for road disease detection, the detection objects are often cracks, holes and other
 categories. Unlike common object detection, there is greater similarity between road diseases and
 the road disease category is more 
 irregular in shape,as shown in Fig.\ref{fig_introduction_1},thus the construction and classification of road disease features 
 are more difficult. In addition, the regional positioning of road diseases such as cracks is 
 also more difficult. In order to make the model more accurately identify the target, we need 
 to better model the object features. Based on this motivation, we believe that contrastive 
 learning will be a good solution. The concept of contrastive learning was proposed as early as 
2006 by \cite{ref22}.The author pointed out that similar points can maintain their similarity when 
 projected into low-dimensional space. In 2020, Google released SimCLR \cite{ref23}, which uses the 
 idea of comparative learning to train the representation of data. \cite{ref24} proposed prototype comparative learning, combined with
the idea of clustering. \cite{ref25} compared cross entropy loss and contrast loss, and 
design new loss function to obtain a more 
balanced representation space. \cite{ref26} introduce virtual labels
and use interpolation to improve  the quality of 
contrast representation learning. \cite{ref27} 
study the hard negative samples  to make the different classes more 
distinguishable. \cite{ref28} proposed that the 
contrast learning process can be optimized by decoupling the amount of learning. \cite{ref29} proved 
that the contrast learning method learns similar information between classes, which may lead to 
the loss of some features that are not similar between classes but are related to tasks.
\par
In this paper,due to the issues mentioned above,we propose LDBFSS(Latent Domain Background Feature Separation and Suppression) network, which could performs 
the separation and suppression of background information without domain supervision 
and the contrastive enhancement of object features. Aiming at the problem that 
background information will interfere with detection in road disease detection,we 
study the domain generalization method and innovatively designed a latent domain discovery
module and a domain adversarial learning module in the network.
The latent domain discovery module separates object features and domain features 
through convolutional feature filtering, and measure the domain similarity of object features 
according to unsupervised clustering on domain features and obtain pseudo domain labels. 
Samples with the same pseudo domain label are considered to have similar background information. 
The domain adversarial learning module takes object features and 
pseudo domain labels as input, and uses the pseudo domain labels to guide the 
optimization training of domain classification. We introduce a gradient reverse 
layer in the domain adversarial learning module to reverse the gradient of domain 
classification loss to achieve adversarial training between the feature extraction 
layer, latent domain discovery module and domain discriminator. During the training 
process,domain discriminator continuously improves the domain classification ability, 
forcing the feature extraction layer and latent domain discovery module to fully 
extract discriminative object features by seperating and suppressing domain features, 
and this in return can provide more 
accurate guidance for the domain discriminator, so that the network can achieve 
progressive enhancement in a adversarial training mode.
In addition, for the problem that road diseases are relatively similar, we 
construct a object-level contrastive learning module, which takes object features as 
input and learns a better disease representation in the latent space by building 
positive sample pairs and negative sample pairs and narrowing 
intra-class features and pulling apart inter-class features. Considering the imbalance
issue in dataset,we design a k-instance contrastive loss which could prevent classes 
with a large amount from dominating the optimization process of contrastive learning 
by balancing the number of positive and negative sample pairs. 
\par
We train and test our method on GRDDC2020 \cite{ref21} and CNRDD \cite{ref20} datasets, and the experimental results 
prove the effectiveness of our method. As far as we know, we first studied the application of 
combination of one-stage object detection model and contrastive learning method in road disease detection.
Besides we first discussed the domain generalization issue in road disease detection and 
extend it to the unsupervised paradigm.Our contributions can be summarized as follows:
\begin{itemize}
    \item[$\bullet$] We first study the domain generalization problem in road disease 
    detection and propose a method for background information separation and suppression  
    in the case of domain-free supervision. We construct a latent domain discovery module 
    to separate features and mine pseudo domain labels in the unsupervised pattern, 
    and construct a domain adversarial learning module to realize the adversarial 
    training of feature extraction layer, latent domain discovery module and domain 
    discriminator, which could continuously enhance the model during the training process.
    \item[$\bullet$]We introduce the contrastive learning method into road disease 
    detection, and through the object-level contrastive learning module, the object 
features are narrowed within the class and distanced between the classes, which 
    strengthens the model's learning of discriminative features of the road disease 
    category. Furthermore, we design a k-instance contrastive loss, which optimizes 
    the contrastive learning process by balancing positive and negative sample pairs.
    \item[$\bullet$] Based on proposed methods, we design a LDBFSS(Latent Domain Background
    Feature Separation and Suppression) network, which is combined with the one-stage object detection 
    model YOLOv5. Experimental results on CNRDD and GRDDC datasets demonstrate the
    effectiveness of our proposed method,in which latent domain discovery module and 
    domain adversarial learning module can reduce the interference of similiar domain 
    features and object-level contrastive learning module can increase the 
    distinguishability of disease representation.
\end{itemize}
\par
The structure of this paper is as follows : In the second section, we introduce the related work
 of the research, and propose our method in the third section. In the fourth section, we show 
 the setting and results of the experiment, and summarize the full text in the fifth section.

\section{RELATED WORK}
In this section, we briefly review the related work of our research.
\subsection{Road Disease Detection}
Traditional road disease detection mainly relies on some basic image features to detect the 
location and category of diseases. For example, \cite{ref10} combines two-dimensional and 
three-dimensional images to detect through visual and spatial geometric information. 
\cite{ref11} combined with directional gradient histogram, using gradient information to achieve 
detection. \cite{ref12} used the local binary pattern to detect the classification of pavement cracks 
in the non-overlapping grid. \cite{ref13} proposed GLBP on this basis, reducing the impact of 
environmental factors on disease detection. Although these traditional methods show the 
effectiveness on some datasets, more complex modeling is needed in the face of more complex road
 disease detection scenarios.
\par
With the development of machine learning, the combination of machine learning and road disease 
detection has attracted more and more attention. \cite{ref14} used EfficientNet \cite{ref36} as the feature 
extraction backbone, and  used BiFPN to achieve multi-scale feature fusion. In \cite{ref15}, 
the author compares the performance of different models based on a variety of feature 
extraction backbones such as CSPDarknet53, Hourglass-104, EfficientNet.In \cite{ref16} the author 
detected cracks under different weather, lighting and other environmental conditions. \cite{ref17} proposed
 to use generative adversarial 
networks and Poisson mixing to generate synthetic images and
 experimental results show that using synthetic images for data enhancement can improve the 
 effect of the detector. In \cite{ref18}, the author selected the new backbone feature extraction 
 network MobileNetV3 to replace the basic network of YOLOv5, reducing the number and size of the
  model parameters. At the same time, the coordinate attention lightweight attention module is 
  introduced to help the network locate the target more accurately and the KMeans clustering 
  algorithm is used to filter the prior frames. Based on the difference between the training 
  dataset and the real test dataset,in \cite{ref19} the unsupervised domain adaptation is introduced 
  , and the model learns the domain-independent feature 
  representation by means of adversarial training, so as to realize the cross-domain domain 
  adaptive disease detection. \cite{ref20} Through prior observation, image processing technology is 
  used to highlight the location of the disease and serve as an attention prompt to guide the 
  model to focus on the disease area, achieving better positioning and feature extraction. In 
  summary, the road disease detection method based on machine learning can handle more complex 
  scenes and has higher detection accuracy. Based on this, our work also studies the innovative 
  application of machine learning technology in the field of road disease detection.
\subsection{Domain Generalization}
Domain generalization aims to use the training data of single or multiple source domains to make 
the model learn domain-independent features, so that it can adapt to the target data of 
different domains. \cite{ref42} used adversarial learning and metric learning to achieve domain 
adaptation, \cite{ref43} considered the influence of noise features unrelated to the target domain on 
domain alignment, and purified the features.\cite{ref44} studied the domain adaptation problem on 
multimodality. \cite{ref45} used domain labels to train domain classifiers and used this training 
model to obtain image representations without domain information. In \cite{ref46}, the author used 
several pre-training models and used ensemble learning to obtain the optimal results. \cite{ref47} 
proposed that classifiers tend to remember the discriminative features of the observed data, 
and designed a classifier ensemble method. \cite{ref48} found an effective training sample weighting 
method to make the model obtain better generalization performance. \cite{ref49} proposed the DANN 
model, which combines domain classifier and feature extractor adversarial training, so that the 
feature extractor extracts domain invariant features. The domain generalization technology can 
deal with the problem of multi-source domain in training data. Road diseases may be sampled from 
different environments, and the same type of disease samples may contain different domain 
information. It is of great value to study the domain difference in road disease detection. 
However, there is a lack of research in this area. This paper will combine the road disease 
detection problem to study the contribution of domain generalization domain to disease detection.
\subsection{Contrastive Learning}
The concept of contrast learning was proposed as early as 2006. \cite{ref22} proposed to 
learn a global uniform nonlinear function that can uniformly map data onto the output manifold 
by learning invariant mappings for dimensionality reduction. The learning is only related to the
neighbor relationship without any distance measure in the input space, and gives the form of 
the earliest contrast loss function. In 2020, Google 's SimCLR \cite{ref23} method uses the idea of
contrastive learning to train the classification model by constructing a contrast loss instead
of the original cross-entropy loss. The unsupervised pattern has achieved the best results in
the classification problem of ImageNet, which proves the feasibility of the contrastive 
learning method in obtaining good feature representation. \cite{ref24} proposed the prototype 
contrast learning PCL,which introduced the prototype as a potential variable, 
constrained the sample features to approach the prototype features, and proposed the 
prototype loss function combined with the contrast loss. \cite{ref25} pointed out that most of the
existing self-supervised learning ( SSL ) methods use artificially balanced datasets ( such
as ImageNet ) to train the representation model. The author proposes K-positive contrast learning to 
generate a
more balanced feature space by balancing the number of positive samples. From the 
perspective of data enhancement, \cite{ref26} introduce virtual labels and perform interpolation 
in the sample space and label space to provide more enhanced data during the training process. 
\cite{ref27} denotes that similar to metric learning,contrastive learning  
benefits from hard negative samples, and the key challenge of mining hard negative samples is 
that the comparison method must remain unsupervised, where samples similar to positive samples 
cannot be directly extracted. Based on this, the author developed a new method for selecting 
difficult negative samples, which can customize the difficulty of negative samples, and proved 
the effectiveness of the method through several downstream tasks. \cite{ref28} theoretically deduced 
the loss function of InfoNCE \cite{ref37} and found that there was a 
significant negative-positive coupling ( NPC ) effect in the loss, which made the learning of 
the model vulnerable to the batch size. This paper proposes the decoupling contrast learning 
( DCL ) loss, which removes the positive term from the denominator and improves 
the learning efficiency. \cite{ref29} proposed that contrast learning tends to learn information 
shared between features, and easily ignores non-shared but task-related information. 
Therefore, it is proposed to increase the mutual information between representation and input, 
so that the model learns non-shared task-related information. \cite{ref30} proposed a more robust 
loss function for the noise pairs that may occur when constructing the positive pair.%
\par
The work mentioned above follows the original paradigm and applies contrastive learning to 
unsupervised representation learning. Considering the superiority of contrastive learning in 
modeling features, some researchers introduce it into supervised learning. \cite{ref31} 
proposed supervised contrast loss, and experiments show the 
consistent superiority compared to cross entropy loss. \cite{ref32} analyzed the cross 
entropy loss and the supervised contrast loss, and provided theoretical verification. \cite{ref33} 
discussed the possible collapse problem in contrastive learning, and improved the robustness of 
the model by enhancing the difference loss within the construction class through data. 
Aiming at the problem of long tail data distribution, a target contrast learning is proposed to 
improve the uniformity of minority feature distribution by making the preset class center evenly 
distributed in the feature space as much as possible\cite{ref34}. In summary, although the research on 
contrastive learning has fully demonstrated its potential, there is still a lack of relevant research 
in road disease detection. This paper will introduce contrastive learning 
methods to enhance the performance of one-stage object detection models in road disease detection.

\section{METHODOLOGY}
\begin{figure*}[!t]
    \centering
    \includegraphics[width=5in]{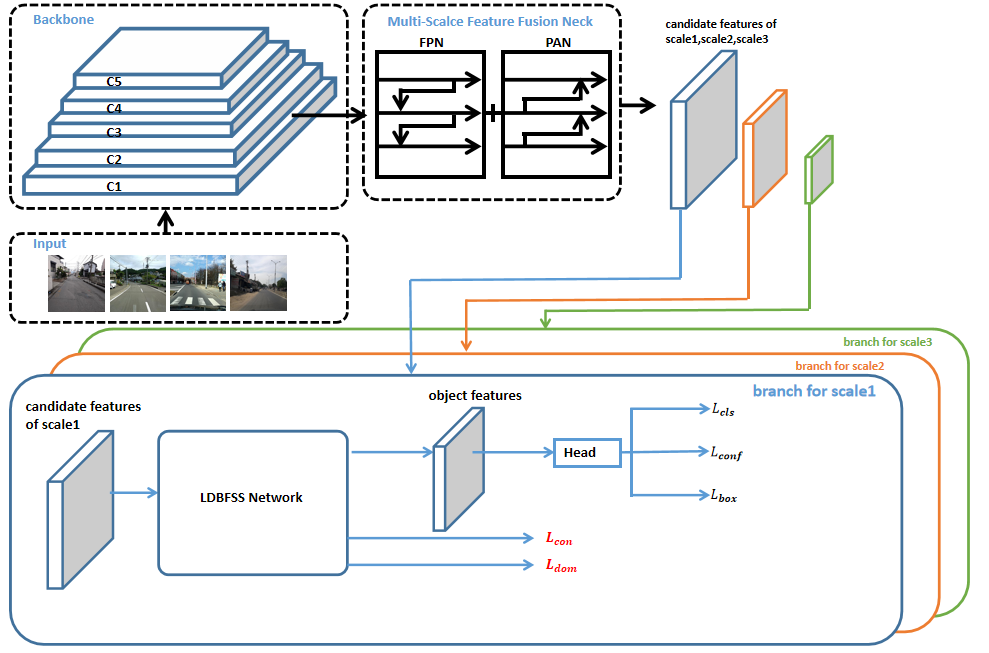}
    \caption{The overall architecture of our model.}
    \label{net1}
\end{figure*}

In this section, we discuss the implementation details of our detection model. The overall structure of the model is 
shown in Fig.\ref{net1}. We use YOLOv5 as the baseline model,the input image is  fed into backbone network and feature 
fusion neck,and the feature maps of three scales are output.A prediction branch is constructed on each scale 
for training,in which predict head takes candidate features as input to predict classification and regression results.
On the basis of the baseline model, 
we additionally construct a LDBFSS(Latent Domain Background Feature Separation and Suppression) network on the prediction branch of each scale.Our network takes the candidate features 
on the current scale  as input and outputs object features with adaptive domain features separation
and suppression and contrastive 
enhancement to predict head for better prediction, at the same time it calculates the 
contrastive loss and domain discriminative loss, which will combine with the predictive loss of the YOLOv5 model 
for optimization to achieve better detection results.

\begin{figure*}[!t]
    \centering
    \includegraphics[width=5in]{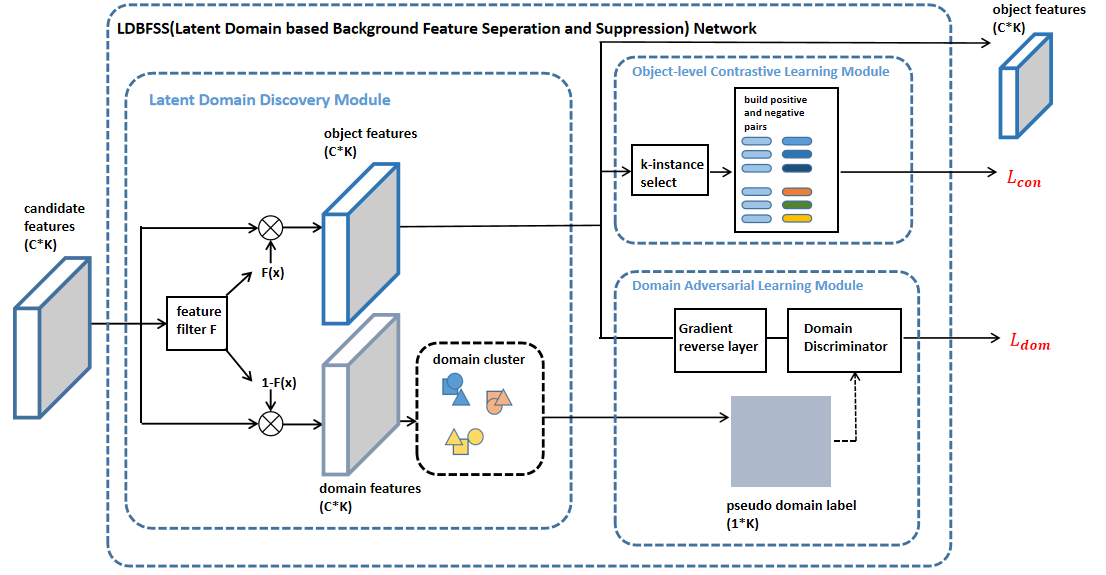}
    \caption{Architecture of LDBFSS network.}
    \label{net2}
\end{figure*}

The structure of LDBFSS(Latent Domain Background Feature Separation and Suppression) network is shown in Fig.\ref{net2},which includes a latent domain discovery module, a domain 
adversarial learning module and a object-level contrastive learning module. 
The latent domain discovery module takes candidate features as input to calculate the filter factor through 
the convolutional feature filter and then perform dot product between filter factor and candidate features as 
feature selecting to separate input features,which obtains domain-compression enhanced object features that 
adaptively reduce the 
interference of background information and domain-related features used for clustering to decide 
pseudo domain labels and realize domain division guidance without domain supervision.
The domain adversarial learning module inputs the object features into domain discriminator, and reduces the 
discriminative loss through the guidance of pseudo domain label. At the same time, we add a gradient reverse 
layer to invert the gradient of discriminative loss, so that the feature extraction layer and the latent 
domain discovery module can be trained with domain discriminator in a adversarial pattern. As training 
progresses, the domain discriminator can predict the domain labels of input features better, forcing 
the feature extraction network and letent  domain discovery module to effectively extract domain-independent object 
features to confuse the domain discriminator.As domain-independent object feature extraction ability improved, 
better domain feature separation is achieved, which in return further trains to obtain a 
better domain discriminator, and realizes the continuous learning enhancement of the module in the adversarial 
training mode.
Object-level contrastive learning module takes object features as input to build positive and negative sample 
pairs,and further obtains more discriminative feature representations by 
narrowing positive pairs features and pulling away negative pairs features. In subsections,we will introduce 
each module in detail.

\subsection{Latent Domain Discovery Module}
For the GRDDC dataset, there is a big difference between the road pictures from the Czech 
Republic and Japan and those from India, and ignoring the background environment information 
is likely to interfere with model training,while previous studies did not 
fully consider the impact of domain differences. 
We design latent domain discovery module, mainly considering two aspects. First, 
in common situations, the collected road disease data often does not contain domain labels, 
and domain labels are not available at all time. Second, domaining data by city source is 
not necessarily the best way. The pictures of 
different cities may have similar background environments, and the pictures of the same city 
may also have large differences. Detecting based on the actual road conditions such as road type, 
road humidity and etc. may have better results, while artificially subdividing and labeling road 
conditions will consume a lot of energy and time. Based on above considerations, this 
paper studies the laten domain discovery method for multi-source data without domain labels.
Specifically,for the candidate features,we use convolutional network as feature filter to learn 
filter factor,and then perform dot product with candidate features as feature selecting to separate 
object features and domain features.From this step,we get object feature which is more suitable for 
prediction,in which the domain-specific information is alleviated.
Then we perform hierarchical clustering on domain features,which could learn a partition automatically
within unsupervised manner.We build pseudo domain label based on clustering result,which could provide
guidance of domain classification correspond to object features.The formula is as follows,$\sigma$ represents Sigmoid function and f represents
convolution neural network for computing filter factor.
    \begin{equation}
        \label{equation6}
        \begin{split}
        x_{obj}=x_i \cdot \sigma (f(x_i))
        \end{split}
    \end{equation}
    \begin{equation}
        \label{equation7}
        \begin{split}
        x_{dom}=x_i \cdot (1- \sigma (f(x_i)) )
        \end{split}
    \end{equation}

\subsection{Domain Adversarial Learning Module}
In order to enable latent domain discovery module to better separate object features and domain features,
inspired by \cite{ref49}, we propose  domain adversarial learning module,in which a 
 discriminator is constructed to classify object features. 
The domain discriminator needs to distinguish which domain the 
feature comes from and optimizes the discriminative loss under the guidance of pseudo domain label,
which guaranteed the continues enhancement of 
discriminative ability as the training processes.
We also add a gradient reverse layer,which reverse the gradient calculated by discriminative loss during
back propogation,thus the feature exrtactor and former latent domain discovery module needs to provide 
domain-independent features that confuse the domain discriminator. 
The continuous learning of the discriminator forces the feature extractor and latent domain discovery 
module to fully extract domain-independent object features, which at the same time obtains better separated
domain features and in return provides 
more accurate learning guidance for the domain discriminator. Through the mode of adversarial learning, we can 
continuously learn to enhance the model during the training process.
Adversarial loss is shown as follows,N is the number of object,
k is the number of domain,E and D represent the feature extractor and discriminator,$y_i$ represents 
the ground truth domain label of sample i,$P_{i,k}$
represents the probability that target i predicts belonging to domain k.
\par

    \begin{equation}
        \label{equation4}
        \begin{split}
        L_{dom}=-\sum_{i=0}^{N}\sum_{d=0}^{K}\mathbb{I} \{y_i=d\}log(P_{i,k})
        \end{split}
    \end{equation}

    \begin{equation}
        \label{equation5}
        \begin{split}
        \theta =arg max_{E}min_D L_{dom}
        \end{split}
    \end{equation}

\subsection{Object-level Contrastive Learning Module}
Considering that diseases classes are realtively diverse and have large similarity, we need 
to extract more discriminative features for the disease category, and most of the existing 
methods simply use a classifier based on the discriminant boundary to guide the category 
learning of the model, which is obviously not enough to complete this task. We therefore 
introduce the contrastive learning method \cite{ref31} to learn more robust feature representations by 
explicitly modeling intra-class similarities and inter-class differences.
\par
In order to implement the introduction of contrastive learning, we design the object-level
contrastive learning module. This module accepts the object feature from latent domain 
discovery module,which has less background information, as input and perform k-instance selected
to randomly choose k samples from each class for balanced training.Then we build positive pairs
and negative pairs based on class label and measure the similarity between sample pairs.We define 
the 
optimization objective,which is shown as (\ref{equation1}),prompts the model to draw closer the 
positive pairs features and pull away negative pairs features and this would makes the object 
features more discriminative and suitable for predicting.
\par
\begin{small}
    \begin{equation}
        \label{equation1}
        \begin{split}
        &\varTheta _{obj}={arg max}_{\varTheta} \sum_{i,j,k\in {\{1,2,...,N\}}and i\neq j\neq k}\mathbb{I} \{y_i = y_j\} \cdot L_d(x_i,x_j) \\ 
        & - \mathbb{I} \{y_i \neq y_j\} \cdot L_d(x_i,x_j) 
        \end{split}
    \end{equation}
    \end{small}
\par
In the supervised context, contrastive learning can use label information to supervise the 
distribution of features and learn more discriminative representations. Inspired by the 
supervised contrastive classification loss \cite{ref31}, we propose a k-instance supervised  
contrastive loss applied to contrastive learning of object features. For an input batch of 
data, we construct the k-instance supervised contrastive loss function $L_{k-instance}$ :
\par
\begin{small}
    \begin{equation}
        \label{equation2}
        \begin{split}
        L_{k-instance}=\frac{1}{N}\sum_{i}^{N}L_{x_i} 
        \end{split}
    \end{equation}
\end{small}
\begin{small}
    \begin{equation}
        \label{equation3}
        \begin{split}
        L_{x_i}=\frac{1}{K+1}\sum_{x_j^+\in X_{i,K}^+}-log(\frac{exp(x_i\cdot x_j^+)}{\sum_{x_k\in x_{i,K}}exp(x_i\cdot x_k)})
        \end{split}
    \end{equation}
\end{small}
N is the number of candidate object in one batch,$X_{i,K}$ represents the object features 
except $x_i$,$X_{i,K}^+$ represents object features share the same class with $x_i$.Specifically,
We randomly sample K-instances within a batch, and compute a contrastive loss between the 
 sample pairs.The original supervised comparison loss \cite{ref31} uses all the same kind of data 
to construct positive sample pairs. When the number of categories in the data is uneven, 
the categories with more instances are decisive for representation learning, which is 
likely to lead to suboptimal results,in contrast our method can better balance the 
learning of different classes through random K-instance sampling, resulting in a more 
uniform and discriminative feature distribution.Similar to our work is \cite{ref25}, but this method 
only adopts a similar method for positive sample pairs, but ignores that the 
value of negative sample pairs in the loss function will also have an impact on the learning 
of the current class. In addition, in the contrastive loss function, the target features are 
normalized first and then dot product, which is equivalent to calculating the cosine 
similarity between feature vectors, which only measures the similarity 
in direction between two vectors. Starting from the fact that the values of different 
channels in the vector correspond to the activation responses of specific features, we 
normalize within the batch and replace the cosine similarity with the dot product similarity. 
Experimental results show that this method generates better representations with the batch 
and optimizes the contrastive learning process.

\subsection{Training Target}
The overall loss of our model consists of classification loss $L_{cls}$,bounding box regression 
loss $L_{reg}$,confidence loss $L_{conf}$,domain classification loss $L_{dom}$ and contrastive 
loss $L_{con}$.Among them,classification loss $L_{cls}$ ,confidence loss $L_{conf}$ and 
domain classification loss $L_{dom}$ adopt the cross entropy loss,regression loss $L_{reg}$ 
adopts CIoU loss and contrastive loss $L_{con}$ compute the k-instance supervised contrastive 
loss.Overall loss is shown as follows,$\lambda$ is set to 0.5 adn $\alpha $ is set to 0.1 to
balance the training.
\begin{small}
    \begin{equation}
        \label{equation8}
        \begin{split}
        L=L_{cls}+L_{reg}+L_{conf}+ \lambda \cdot L_{con} +\alpha \cdot L_{dom}
        \end{split}
    \end{equation}
\end{small}

\section{EXPERIMENTS}
We experimented the proposed method on two road disease recognition datasets 
GRDDC2020 \cite{ref21} and CNRDD \cite{ref20}. In addition, we carried out ablation experiments to verify 
the performance of each part of the model. In this section, we first introduce the 
experimental setup, second we present and discuss the results on the road disease detection 
dataset, and finally we present the results of the ablation experiments.
\subsection{Training Setting}
Our experiments are based on the PyTorch framework and implemented on the CentOS system. 
Four Tesla V00 graphics cards are used in the experiment, the batch size is 64, and the 
epoch is 100. We use the SGD optimizer, the momentum is set to 0.937, and the weight decay 
is 5e-4.
\subsection{Dataset Setting}
{\bf{GRDDC2020}} \cite{ref21}. The GRDDC dataset contains a training set and two test sets,test1 and 
test2, including 9 disease categories ('d00', 'd01', 'd0w0', 'd10', 'd11', 'd20', 'd40' , 
'd43', 'd44', 'd50'), collected from three different countries (Czech Republic, Japan and 
India). The training set contains 21041 pictures, test1 contains 2631 pictures, test2 
contains 2664 pictures, and only the training set contains labeled data. We follow the data 
pattern of GRDDC2020, conduct training on four categories: longitudinal cracks, transverse 
cracks,aligator cracks and potholes (d00, d10, d20, d40), and submit the results to the official 
platform (https://rdd2020. sekilab.global) for testing. In detail, we used one-tenth of the 
data in the training set as the validation set and submitted the model that performed best 
on the validation set. The experimental results are evaluated using the official indicator F1.
\par
{\bf{CNRDD}} \cite{ref20}. CNRDD is a road disease detection dataset in China proposed in \cite{ref20}, 
which contains 4295 pictures of road asphalt pavement, including 3022 training sets and 1273 
test sets. The data disease types are divided into 8 kinds of Crack, Longitudinal Crack, 
Lateral Crack, Subsidence, Rutting, Pothole, Looseness, and Strengthening. This experiment 
is conducted according to the division of training and test sets given by the dataset 
provider, and the experimental results are evaluated on mAP and F1.

\subsection{Road Disease Detection}
We compared the performance of the proposed method with several disease detection 
methods, such as: SUTPC\cite{ref50}, Dongjuns\cite{ref51}, DD-VISION\cite{ref52}, SIS Lab\cite{ref53}, 
IMSC\cite{ref54}, E-LAB\cite{ref55}, IDVL\cite{ref56}, BDASL\cite{ref57}, CS17\cite{ref58},
AIRS-CSR\cite{ref59}, RICS\cite{ref60}, tian$\underline{~}$mu\cite{ref61}.All methods are trained 
on the GRDDC2020 training dataset, and 
then calculate the F1 score on the test1 and test2 test sets,as shown in Table \ref{tab1}. In the 
right column of the 
table, Ensemble means that the model uses YOLOv5, Faster R-CNN and other models for 
ensemble learning. Single means that a single model method is used. Our method only uses 
YOLOv5, so it is divided into a single model method. From the Table \ref{tab1}, we can observe that the 
average performance of the ensemble learning method is better than that of the single 
model method. For example, ISMC\cite{ref54} uses the Faster R-CNN and YOLO models. The SIS method\cite{ref53} 
employs YOLO models of different resolutions. DD-VISION\cite{ref52} integrates Faster-RCNN, ResNeXt-101, 
HR-Net, CBNet, ResNet-50 and other methods. The ensemble learning method improves the 
accuracy of road damage detection by integrating different features, but it also greatly 
increases the training overhead and deployment difficulty. In the single-model method, our 
model achieved the best results, with F1 reaching 62.1\% on test1 and 61.1\% on test2, 
which is 4\% and 3.5\% higher than the best tian$\underline{~}$mu\cite{ref61} in the single model.
\par
We also conduct experiments on the CNRDD dataset, comparing the test performance of the 
proposed method with YOLOv5, Faster R-CNN, Fcos\cite{ref62}, Dongjuns\cite{ref51}, \cite{ref20} as shown
in Table \ref{tab2}. It 
can be seen that our 
method achieves the best scores in both mAP and F1 score, among which mAP reaches 31.2\%, which 
is 6.4\% higher than the baseline YOLOv5 and 4.6\% higher than the 
previous best method in \cite{ref20}.Our model also achieves a better improvement 
on the F1 score, which is 5.4\% and 2.9 \% higher than the baseline model and 
\cite{ref20}.

\subsection{Ablation Study}

\begin{table}
\begin{center}
\caption{ABLATION STUDY ON GRDDC DATASET}
\label{tab3}
\resizebox{0.5\textwidth}{!}{
\begin{tabular}{ c | c | c | c | c | c | c }
\toprule
\hline
Model & w/w.o               & w/w.o               & w/w.o               & w/w.o               & Test1 F1 & Test2 F1\\
      &contrastive & k-instance    & domain  & latent      &          & \\
      & learning & loss     & adversarial  &  domain       &          & \\
\hline
Ours  &x&x&x&x&53.1&52.4\\
\hline
Ours  &\checkmark&x&x&x&58.8&57.6\\
\hline
Ours  &\checkmark&\checkmark&x&x&60.4&60.1\\
\hline
Ours  &\checkmark&\checkmark&\checkmark&x&60.9&60.8\\
\hline
{\bf{Ours}}  &\checkmark&\checkmark&\checkmark&\checkmark&{\bf{62.1}}&{\bf{61.1}}\\
\hline
\bottomrule
\end{tabular}
}
\end{center}
\end{table}

\begin{table}
\begin{center}
\caption{ABLATION STUDY ON CNRDD DATASET}
\label{tab4}
\resizebox{0.5\textwidth}{!}{
\begin{tabular}{ c | c | c | c | c | c }
\toprule
\hline
Model & w/w.o               & w/w.o               & w/w.o               & w/w.o               & mAP@0.5 \\
    &contrastive & k-instance      & domain  & latent       &           \\
    &learning & loss     & adversarial  &  domain       &           \\
\hline
Ours  &x&x&x&x&24.8\\
\hline
Ours  &\checkmark&x&x&x&28.7\\
\hline
Ours  &\checkmark&\checkmark&x&x&29.3\\
\hline
Ours  &\checkmark&\checkmark&\checkmark&x&29.9\\
\hline
{\bf{Ours}}  &\checkmark&\checkmark&\checkmark&\checkmark&{\bf{31.2}}\\
\hline
\bottomrule

\end{tabular}
}
\end{center}
\end{table}
We conducted ablation experiments on the GRDDC2020 and CNRDD datasets. Table \ref{tab3} and 
Table \ref{tab4} show the 
impact of incrementally adding components over the original YOLOv5 baseline. It can be seen 
that the introduction of contrastive learning training has improved the two datasets of 
the model, among which the F1 score on the GRDDC has increased by 5\%, and the mAP on the 
CNRDD dataset has increased by about 4\%. The introduction of k-instance supervised 
contrastive loss has further improved the model’s experimental indicators. The F1 score 
on GRDDC has increased by 2\%, and the mAP indicator on CNRDD has increased by 0.6\%.Further,
our research of domain adversarial learning and latent domain discovery also improve the detect
performance,which makes about 2\% higher on F1 and mAP.

\subsection{Latent Domain and City Domain}

\begin{table}
    \begin{center}
    \caption{ABLATION STUDY ON CNRDD DATASET}
    \label{tab5}
    \resizebox{0.33\textwidth}{!}{
    \begin{tabular}{ c | c | c }
    \toprule
    \hline
    K    &Test1 F1           & Test2 F1           \\
    \hline
    k=3  &61.5 &60.6\\
    \hline
    k=4  &61.8&60.8\\
    \hline
    {\bf{k=5}}  &{\bf{62.1}}&{\bf{61.1}}\\
    \hline
    k=6  &61.5&60.5\\
    \hline
    \bottomrule
    \end{tabular}
    }
    \end{center}
    \end{table}
For the latent domain discovery module, we conduct related experiments on the 
GRDDC dataset. First, we conducted experiments on the number of clusters k and 
Table \ref{tab5} shows the impact of different numbers of clusters on the results. 
 When the number of clusters k=5, the model achieved the 
best performance. The experimental results prove to a certain extent that it 
is not necessarily the best choice to divide the domain based on the city 
source of the road disease picture and determine the number of domains based 
on it, and our method can obtain a better result by mining the latent domain 
according to the number of clusters k.

\begin{figure}[htbp]
    \begin{minipage}[t]{0.5\linewidth}
        \centering
        \includegraphics[width=\textwidth]{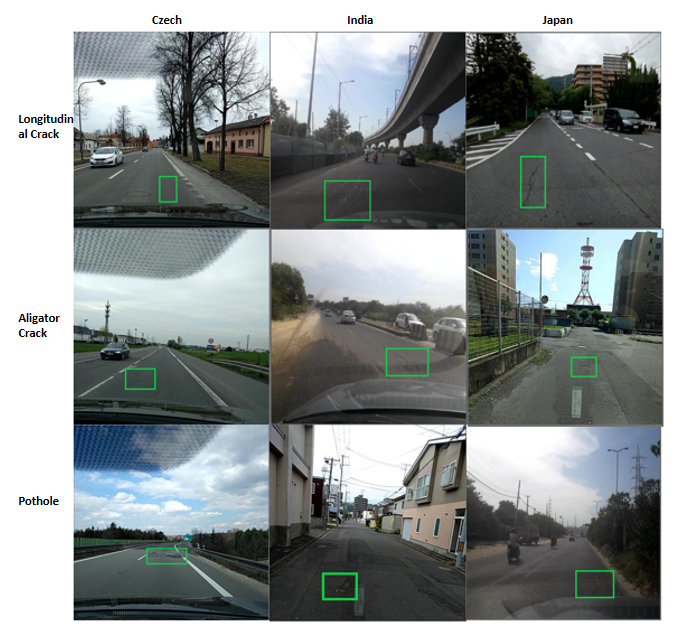}
        \centerline{(a)}
    \end{minipage}%
    \begin{minipage}[t]{0.5\linewidth}
        \centering
        \includegraphics[width=\textwidth]{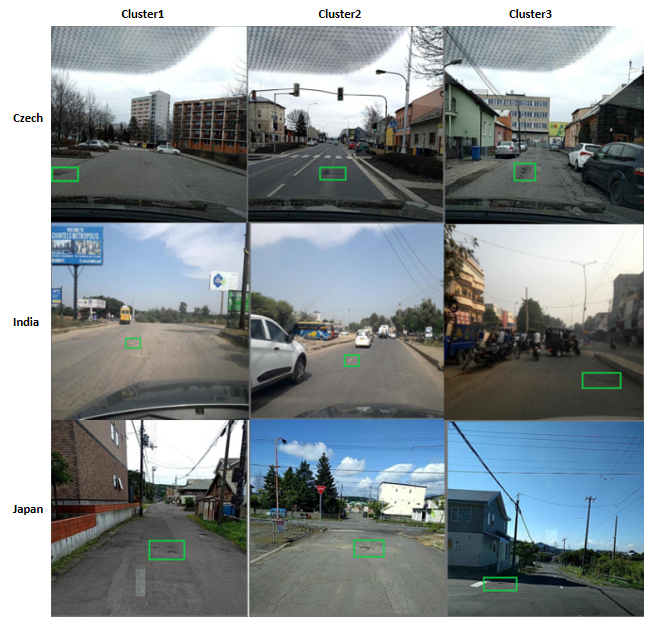}
        \centerline{(b)}
    \end{minipage}
    \caption{Latent domain discovery results.
    (a)Each row shows images clustered by the model into the same cluster.(b)Potholes clustering
    results.}
    \label{fig_latent_domain}
\end{figure}

We analyze the clustering results of the latent domain discovery module, first 
we found that for the same domain cluster of the same disease category, the 
disease targets can come from different cities. As shown in Fig.\ref{fig_latent_domain}(a), 
each row shows images clustered by the model into the same cluster. 
It can be seen from the figure that for the categories of longitudinal cracks, 
aligator cracks, and potholes, there are cases in the dataset where the city source 
labels are different while the environmental background of the disease target 
region is quite similar,thus it is easy to ignore this similarity when 
dividing the domain according to the city label ,which is not reasonable . 
But in our latent domain discovery method, the model can perform adaptive domain 
clustering according to similar backgrounds. For the same disease category, 
the model classifies disease targets from different cities into the same cluster.
\par
Second, we found that the same disease category from the same city was divided 
into different clusters.
Fig.\ref{fig_latent_domain}(b) shows the results of the pothole category.  It 
can be seen from the figure that there are cases in the dataset where the 
source of the city is the same, but the background environment of the disease 
target is significantly different. It is easy to ignore this difference 
if it is divided into the same cluster according to the city label. In our 
latent domain discovery 
method, The model can perform adaptive domain clustering according to different 
backgrounds. For the same disease category, the model classifies disease targets 
from the same city into different clusters.

\subsection{Constrastive Learning}
\begin{figure}[!t]
    \includegraphics[width=0.48\textwidth]{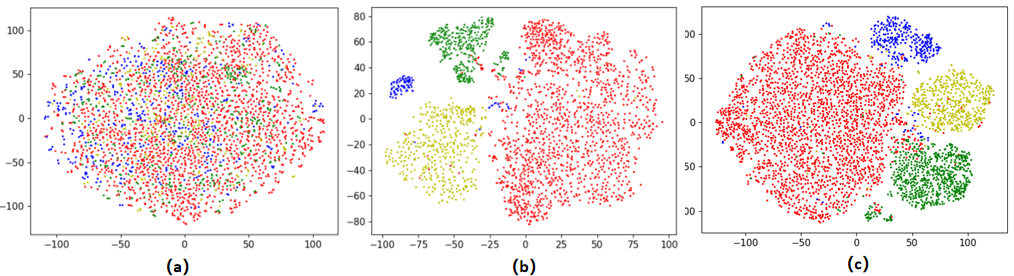}
    \caption{T-SNE visualization of GRDDC validation dateset1 object features,corresponding
    to features before training,features trained with simply introduced contrastive
    method and features extracted from our model.}
    \label{fig4}
\end{figure}
We used the T-SNE method to visualize the target features on the GRDDC2020 
validation set for dimensionality reduction to facilitate our discussion of 
the role of contrastive learning,which is shown in Fig.\ref{fig4}. 
From left to right in the figure shows features before training,features 
trained with simply introduced contrastive
method and features extracted from our model. 
From the figure, we can see that after using the contrastive learning method, 
the model can effectively form clusters of different categories of targets, 
thereby achieving a better discriminant boundary, which demonstrates the help 
of introducing contrastive learning in the road disease detection task. It can 
be seen from the last two figures that our proposed method can form more 
discriminative clusters in the feature space. In addition, it can be seen that 
the boundaries of clusters and the distribution of clusters are more uniform 
in comparison.

\subsection{Analysis of Domain Discovery and Adversarial Learning}
\begin{figure}[htbp]
    \begin{minipage}[t]{0.33\linewidth}
        \centering
        \includegraphics[width=\textwidth]{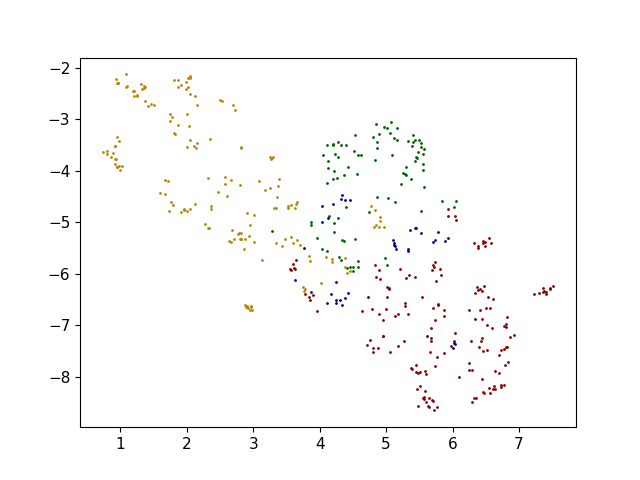}
        \centerline{(a)}      
    \end{minipage}%
    \begin{minipage}[t]{0.33\linewidth}
        \centering
        \includegraphics[width=\textwidth]{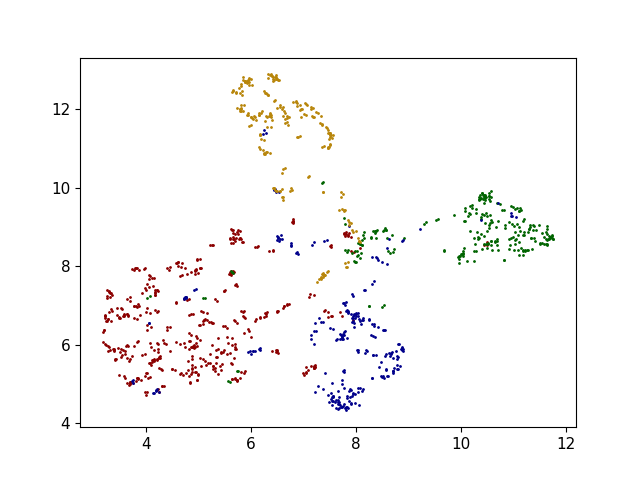}
        \centerline{(b)} 
    \end{minipage}%
    \begin{minipage}[t]{0.33\linewidth}
        \centering
        \includegraphics[width=\textwidth]{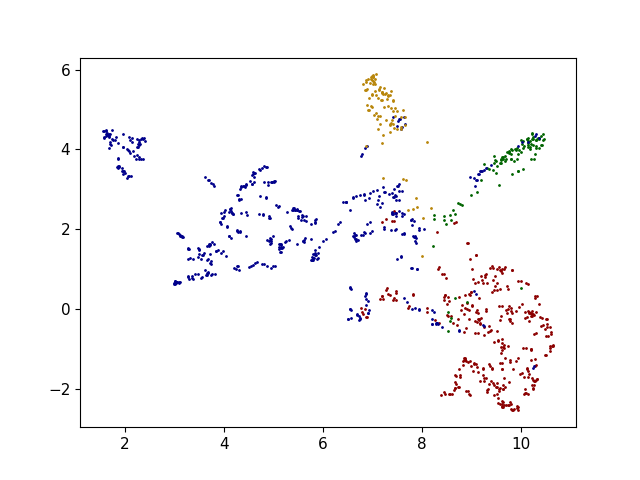}
        \centerline{(c)} 
    \end{minipage}
    \qquad
    \begin{minipage}[t]{0.33\linewidth}
        \centering
        \includegraphics[width=\textwidth]{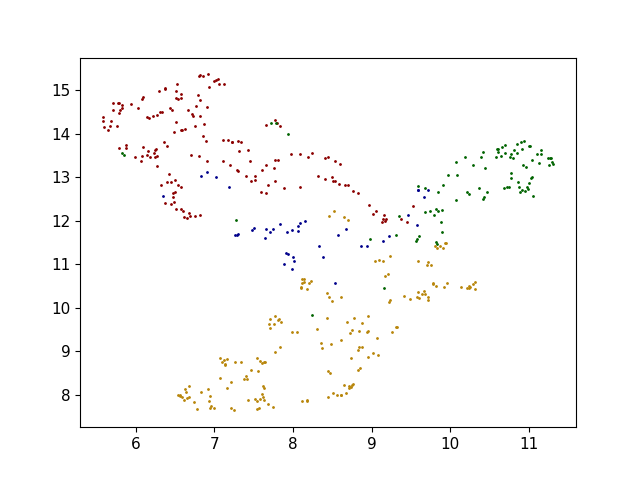}   
        \centerline{(d)}   
    \end{minipage}%
    \begin{minipage}[t]{0.33\linewidth}
        \centering
        \includegraphics[width=\textwidth]{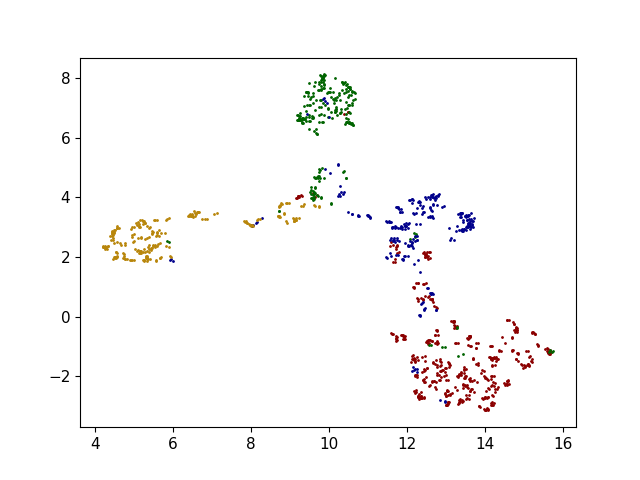}
        \centerline{(e)} 
    \end{minipage}%
    \begin{minipage}[t]{0.33\linewidth}
        \centering
        \includegraphics[width=\textwidth]{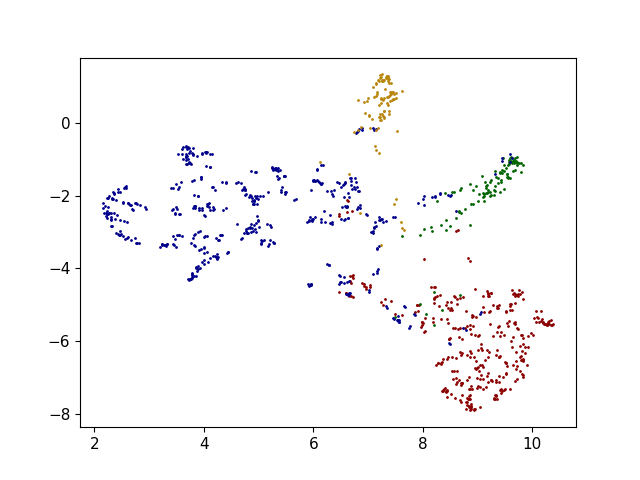}
        \centerline{(f)} 
    \end{minipage}
    \qquad
    \begin{minipage}[t]{0.33\linewidth}
        \centering
        \includegraphics[width=\textwidth]{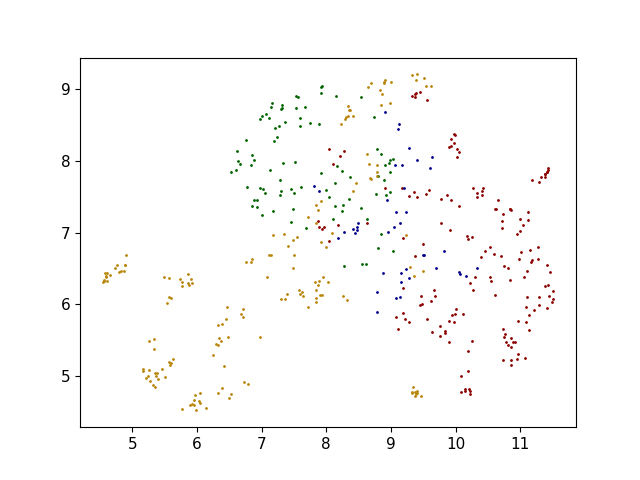} 
        \centerline{(g)}     
    \end{minipage}%
    \begin{minipage}[t]{0.33\linewidth}
        \centering
        \includegraphics[width=\textwidth]{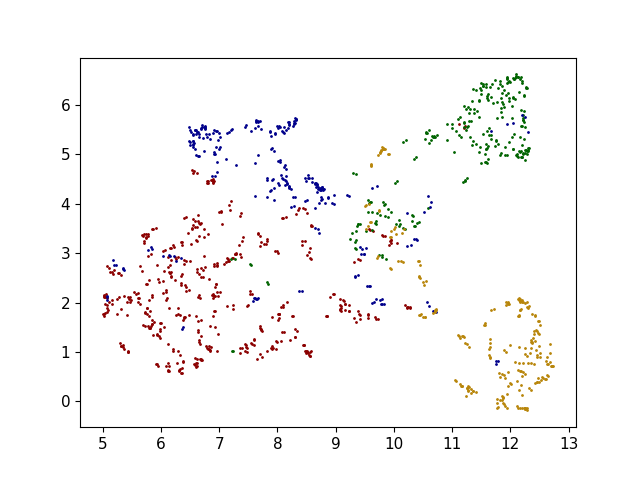}
        \centerline{(h)} 
    \end{minipage}%
    \begin{minipage}[t]{0.33\linewidth}
        \centering
        \includegraphics[width=\textwidth]{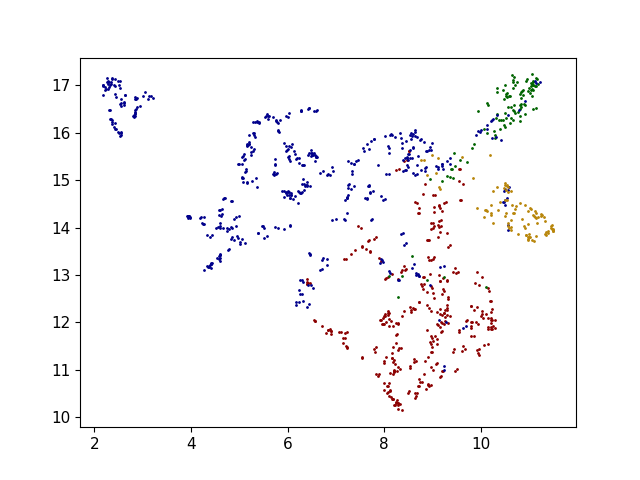}
        \centerline{(i)} 
    \end{minipage}
    \caption{Distribution of candidate features,object features and
    domain features on three scales,colored by class label.
    Each row corresponds to feature type 
    and each column corresponds to scale1,scale2 and scale3.}
    \label{fig_class}
\end{figure}

We used UMAP method to visulize feature informations within one batch from 
country Czech to study the performance of latent domain discovery module
and domain adversarial learning module.As shown in Fig.\ref{fig_class},
we show the feature distribution of candidate features,object features and
domain features on three different scales from large to small with the channel
size 128,256,and 512.First,it can be seen that the candidate feature distribution 
(a)-(c) can be roughly regarded as the combination of the object 
features(d)-(f) and the domain features(g)-(i). 
For features of all scales,as the object features and domain features 
are separated,we observe a better distribution shapes for object features 
and find better distinction between classes,which makes an improvement on 
confusion region in the center.Second,the domain features visualization
shows that there is domain feature difference even between samples from 
same class of Czech,which could be clearly observed from the scale2(h) and 
scale3(i).

\par
Then we use clustering results to color the above distribution,and analyse
the optimization of latent domain discovery module and domain adversarial
learning module to the featues.

\begin{figure}[htbp]
    \begin{minipage}[t]{0.33\linewidth}
        \centering
        \includegraphics[width=\textwidth]{fig6_total_class2.png}
        \centerline{(a)} 
    \end{minipage}%
    \begin{minipage}[t]{0.33\linewidth}
        \centering
        \includegraphics[width=\textwidth]{fig6_domain_class2.png}
        \centerline{(b)} 
    \end{minipage}%
    \begin{minipage}[t]{0.33\linewidth}
        \centering
        \includegraphics[width=\textwidth]{fig6_object_class2.png}
        \centerline{(c)} 
    \end{minipage}%
    \qquad
    \begin{minipage}[t]{0.33\linewidth}
        \centering
        \includegraphics[width=\textwidth]{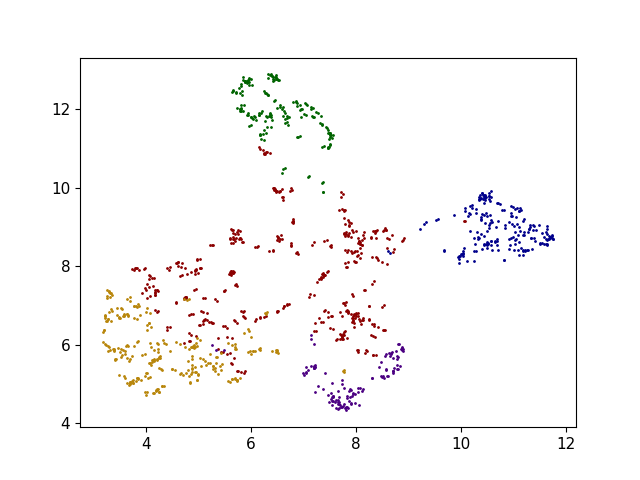}
        \centerline{(d)} 
    \end{minipage}%
    \begin{minipage}[t]{0.33\linewidth}
        \centering
        \includegraphics[width=\textwidth]{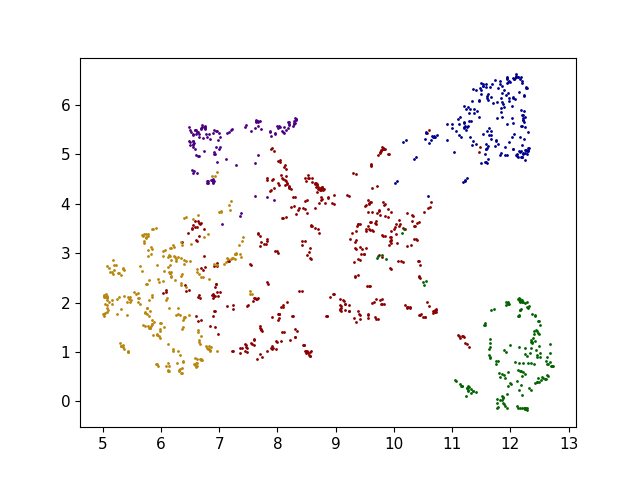}
        \centerline{(e)} 
    \end{minipage}%
    \begin{minipage}[t]{0.33\linewidth}
        \centering
        \includegraphics[width=\textwidth]{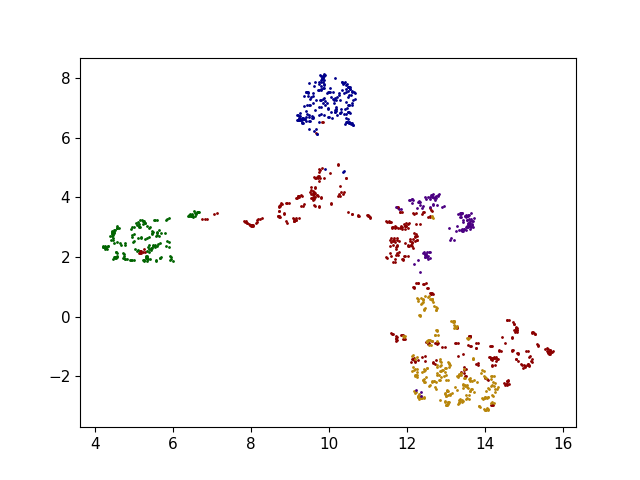}
        \centerline{(f)} 
    \end{minipage}
    \caption{Features distribution on scale2.(a)-(c)candidate features,domain features and
    object features colored by class label.(d)-(f)Same features but colored by clustering
    result.
    }
    \label{fig_case1}
\end{figure}

First,we study the issue that the presence of similar domain information between different 
classes confuses the discrimination.As shown in Fig.\ref{fig_case1}
for scale2's features,we observe a confusion region in the center of candidate features(graph(a)),
indicating that there are similar domain information between these samples from 
different classes.Graph(b) and graph(e) show the domain features extracted by our model and 
the dense mixed zone of domain features from different classes in the center of graph(b)
is assigned with the red color cluster,which will provide the model with the knowledge that 
those samples have same pseudo domain label.With such guidance,domain discriminator takes
corresponding object features from different class as input and needed to extract domain-related
features to learning tht consistency,which would force the model to alleviate domain information
in object features adversarially.And from cluster colored graph(d) and graph(e),we could find
that the dense mixed zone in our obtained feature distribution(graph(b)) is corresponding
to that of candidate features(graph(a)),indicates that our model work by discovering the
latent similar domain features and perform suppression.Graph(c) and graph(f) shows that binary
seperating and suppressing domain features,we obtained better representation.

\par
\begin{figure}[htbp]
    \begin{minipage}[t]{0.33\linewidth}
        \centering
        \includegraphics[width=\textwidth]{fig6_total_class3.png}
        \centerline{(a)} 
    \end{minipage}%
    \begin{minipage}[t]{0.33\linewidth}
        \centering
        \includegraphics[width=\textwidth]{fig6_domain_class3.png}
        \centerline{(b)} 
    \end{minipage}%
    \begin{minipage}[t]{0.33\linewidth}
        \centering
        \includegraphics[width=\textwidth]{fig6_object_class3.png}
        \centerline{(c)} 
    \end{minipage}%
    \qquad
    \begin{minipage}[t]{0.33\linewidth}
        \centering
        \includegraphics[width=\textwidth]{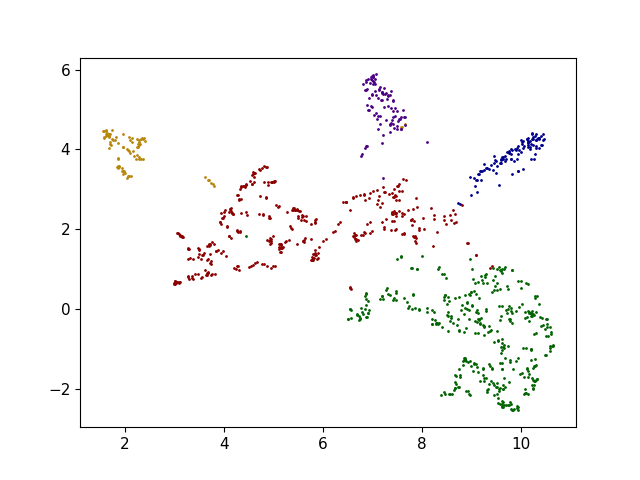}
        \centerline{(d)} 
    \end{minipage}%
    \begin{minipage}[t]{0.33\linewidth}
        \centering
        \includegraphics[width=\textwidth]{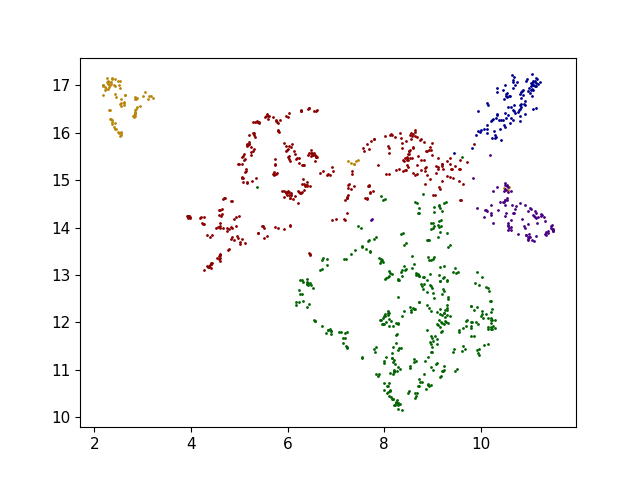}
        \centerline{(e)} 
    \end{minipage}%
    \begin{minipage}[t]{0.33\linewidth}
        \centering
        \includegraphics[width=\textwidth]{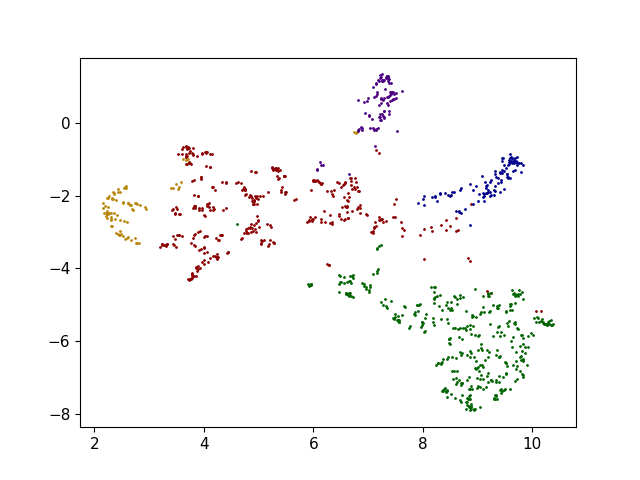}
        \centerline{(f)} 
    \end{minipage}
    \caption{Features distribution on scale3.(a)-(c)candidate features,domain features and
    object features colored by class label.(d)-(f)Same features but colored by clustering
    result.
    }
    \label{fig_case2}
\end{figure}
Then we analyse the influence of clustering numbers k. Fig.\ref{fig_case2} show the Features
distribution on scale3.For candidate features in graph(a),
features of same class can be subdivided into several small clusters,such as the blue color
samples at the left-top region. This might be due 
to the large convolutional area corresponding to the feature points, resulting in more 
diverse domain information.As we mentioned before,this would preserved in domain features(graph(b))
and from clustering result of domain features shown in graph(e),model 
assigns different pseudo labels to these small clusters,takes the blue color samples in graph(a)
as example,which are divided into yellow cluster red cluster and thus may influence the 
learning of mixed-class region due to the limited number of cluster number k=5,thus we 
need to choose a appropriate number for clusters.
On the other side,domain discriminator need to extract domain-related feature from objecct features 
to learn the difference,enabling the adversarial network to compress domain-related information and 
improve representation,as shown in graph(c),where the two part of blue samples are closer by 
domain feature suppression.

\subsection{Inference Detection}
\begin{figure}[!t]
    \includegraphics[width=0.5\textwidth]{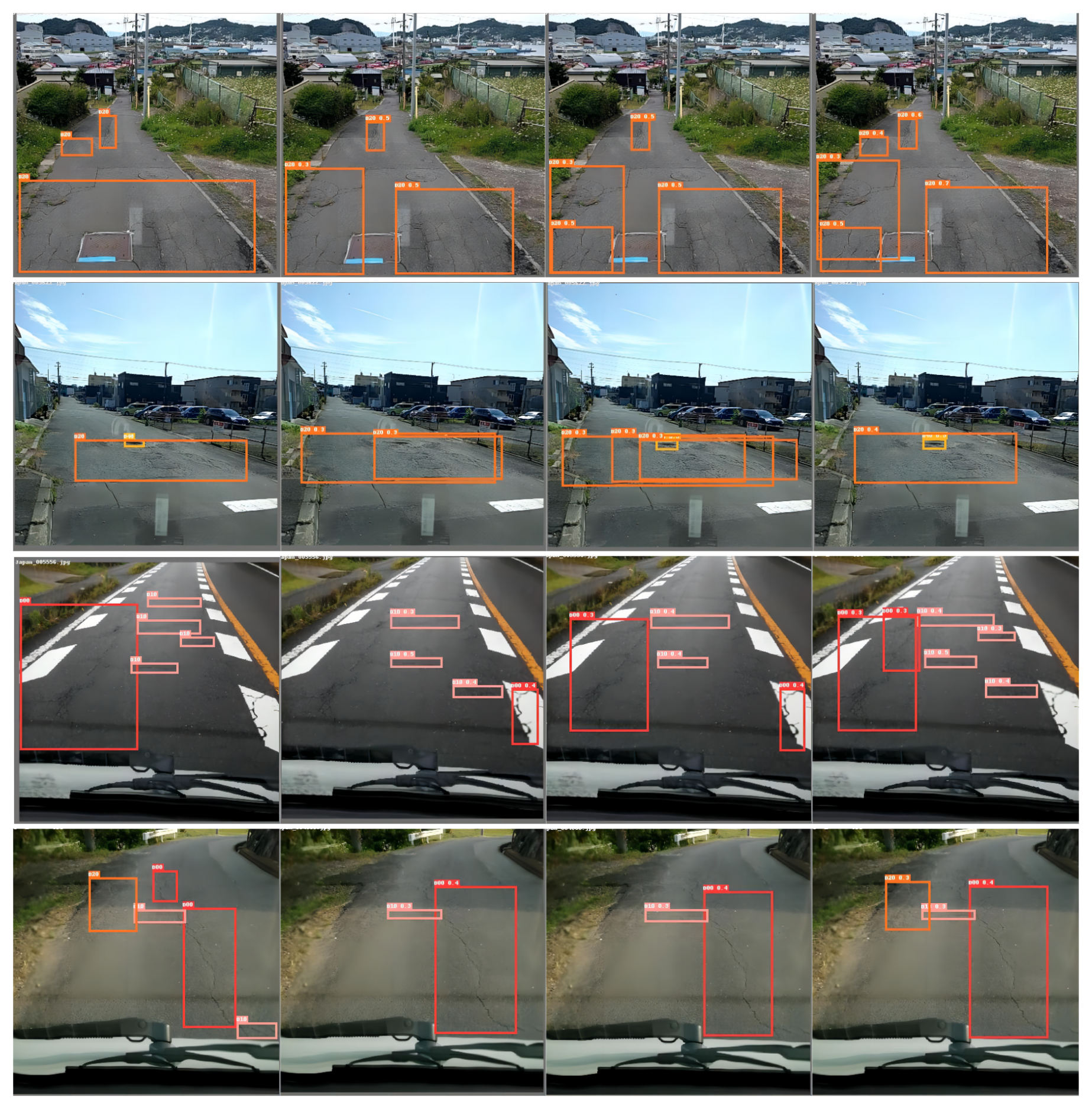}
    \caption{Inference detection results corresponding
    to the GT, YOLOv5 (Baseline) results, detection results of simply
    introduce contrastive learning method, and results of our method 
    from left to right by column}
    \label{fig5}
\end{figure}

Finally, we show different detection performances of several methods on 
GRDDC road images. Fig.\ref{fig5} shows figures of ground truth value, 
YOLOv5 (Baseline) detection results, detection results of simply
 introduce contrastive learning method, and detection results of our model 
from left to right by column . It can be seen that our proposed 
method can achieve better detection results.

\section{CONCLUSION}
In this work, we propose LDBFSS(Latent Domain Background Feature
separation and suppression) method for road disease detection,which could perform feature
separation and suppression to alleviate background information that may interfere
road disease detection.With the introduce of contrastive learning,our LDBFSS method
can extract more discriminative object features.Specifically,we design the LDBFSS network
and combine it with YOLOv5 model to improve detection,which includes three components.
We first design a latent domain discovery module to adaptively separate and suppress background
features by convolutional feature filter and obtain better object features.Background features
are used for clustering to obtain pseudo domain label under domain supervision free scene. 
Then we construct a domain adversarial learning network,
in which a domain discriminator takes object features as input and pseudo domain label as guidance,
learning to classify object features well,while the feature extraction layers and latent domain discovery
module learn to provide object features that could confuse the discriminator,thus learn to alleviate the
domain-related information.The improvement of separation ability of model enable it to provide
better domain features to strengthen the domain discriminator and achieve adversarial training with
domain discriminator.Besides,we construct a object-level contrastive learning module,
which increases intra-class consistency and inter-class discrimination by performing contrastive learning 
on object features. In addition, we consider the the balance issue in training 
data, and propose k-instance contrastive loss function, which 
optimizes the contrastive training by balancing the number of positive pairs and negative pairs. 
We conducted experiments on two public benchmark datasets GRDDC2020 and CNRDD and 
achieved the best results, proving the effectiveness of the method. Our T-SNE and UMAP 
visualization results and inference detection results also show that 
our detection model benefits from our proposed method and achieves better detection results.
\par
In the future, we will continue to study more effective and efficient detection 
algorithms of road disease detection, and further 
explore the combination of latent domain discovery, contrastive learning and road disease  
 detection problems. Finally, we hope that our work can stimulate more 
 researches on road disease detection.

  \vspace{11pt}
  
  \begin{IEEEbiography}
    [{\includegraphics[width=1in,height=1.25in,clip,keepaspectratio]{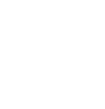}}]
    {JUWU ZHENG}
    received the BE degree from the
    Sun Yat-sen University,
    Guangdong, China, in 2017. He is currently working
    toward the ME degree in the School of Computer 
    Science and Engineering, Sun Yat-sen University. 
    His research interests
    include computer vision and machine learning.
  \end{IEEEbiography}

  \begin{IEEEbiography}
    [{\includegraphics[width=1in,height=1.25in,clip,keepaspectratio]{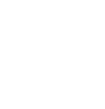}}]
    {JIANGTAO REN}
    received the bachelor's degree in 1998 and the PhD 
    degree in 2003 from Tsinghua University. He is currently an Associate 
    Professor with the School of Computer Science and Engineering, Sun Yat-Sen 
    University. His research interests include data mining and knowledge 
    discovery, machine learning.
  \end{IEEEbiography}

  \vfill

\end{document}